\documentclass[11pt]{article}
\usepackage{ACL2023}

\usepackage{times}
\usepackage{latexsym}

\usepackage[T1]{fontenc}

\usepackage{microtype}
\usepackage{ifthen}
\usepackage{inconsolata}
\usepackage{algorithm}
\usepackage{algorithmic}
\PassOptionsToPackage{table,dvipsnames}{xcolor}
\usepackage{xcolor}
\usepackage{colortbl}
\usepackage{booktabs} 
\usepackage{siunitx} 
\usepackage{amsmath} 
\usepackage{amssymb}

\usepackage{amsmath}
\usepackage{graphicx}
\usepackage{multirow}
\usepackage{multicol}
\usepackage{rotating}
\usepackage[caption=false,font=footnotesize]{subfig}
\usepackage{fancyhdr}
\usepackage{lipsum} 
\usepackage{longtable}

\newcommand{\MyFC}{GraphFC}
\newtheorem{assertion}{Assertion}
\usepackage{listings}
\usepackage{soul}
\definecolor{backcolour}{rgb}{0.95,0.95,0.92}
\definecolor{codegreen}{rgb}{0,0.6,0}
\definecolor{intropurple}{rgb}{133,99,159}
\lstdefinestyle{myStyle}{
    backgroundcolor=\color{backcolour},   
    commentstyle=\color{codegreen},
    basicstyle=\ttfamily\tiny,
    breakatwhitespace=false,         
    breaklines=true,                 
    keepspaces=true,                 
    numbers=none,       
    numbersep=5pt,                  
    showspaces=false,                
    showstringspaces=false,
    showtabs=false,                  
    tabsize=2,
    frame=single
}

\lstdefinestyle{myStyle*}{
    backgroundcolor=\color{white},   
    commentstyle=\color{codegreen},
    basicstyle=\linespread{1.2}\ttfamily\scriptsize,
    breakatwhitespace=false,         
    breaklines=false,                 
    keepspaces=true,                 
    numbers=none,       
    numbersep=5pt,                  
    showspaces=false,                
    showstringspaces=false,
    showtabs=false,                  
    tabsize=2,
    frame=single
}

\title{A Graph-based Verification Framework for Fact-Checking}

\author{
Yani Huang$^{1}$, Richong Zhang$^{1,2}$\thanks{\ \ Corresponding author.}, Zhijie Nie$^{1}$, Junfan Chen$^{3}$, Xuefeng Zhang$^{1}$  \\
$^{1}$CCSE, School of Computer Science and Engineering, Beihang University, Beijing, China \\
$^{2}$Zhongguancun Laboratory, Beijing, China \\
$^{3}$School of Software, Beihang University, Beijing, China \\
\texttt{\{huangyn, zhangxf\}@buaa.edu.cn} \\
\texttt{\{zhangrc, niezj, chenjf\}@act.buaa.edu.cn}
}

\begin{document}
\maketitle
\begin{abstract}
Fact-checking plays a crucial role in combating misinformation. Existing methods using large language models (LLMs) for claim decomposition face two key limitations: (1) \textit{insufficient decomposition}, introducing unnecessary complexity to the verification process, and (2) \textit{ambiguity of mentions}, leading to incorrect verification results. To address these challenges, we suggest introducing a claim graph consisting of triplets to address the insufficient decomposition problem and reduce mention ambiguity through graph structure. Based on this core idea, we propose a graph-based framework, GraphFC, for fact-checking. The framework features three key components: \textit{graph construction}, which builds both claim and evidence graphs; \textit{graph-guided planning}, which prioritizes the triplet verification order; and \textit{graph-guided checking}, which verifies the triples one by one between claim and evidence graphs. Extensive experiments show that GraphFC enables fine-grained decomposition while resolving referential ambiguities through relational constraints, achieving state-of-the-art performance across three datasets.

\end{abstract}

\section{Introduction}\label{sec:intro}

Fact-checking plays a crucial role in detecting misinformation and preventing the spread of rumors. 
The recent emergence of LLMs, which exhibit powerful semantic understanding capabilities, has opened up new potential solutions for fact-checking. Leveraging these LLMs, recent methodologies \cite{pan2023fact, wang2023explainable, zhao2024pacar} have introduced approaches that decompose claims into textual sub-claims. Such LLM-driven claim decomposition simplifies the verification process and allows for more precise identification of errors within the claim.

\begin{figure}
    \centering
    \includegraphics[width=1\linewidth]{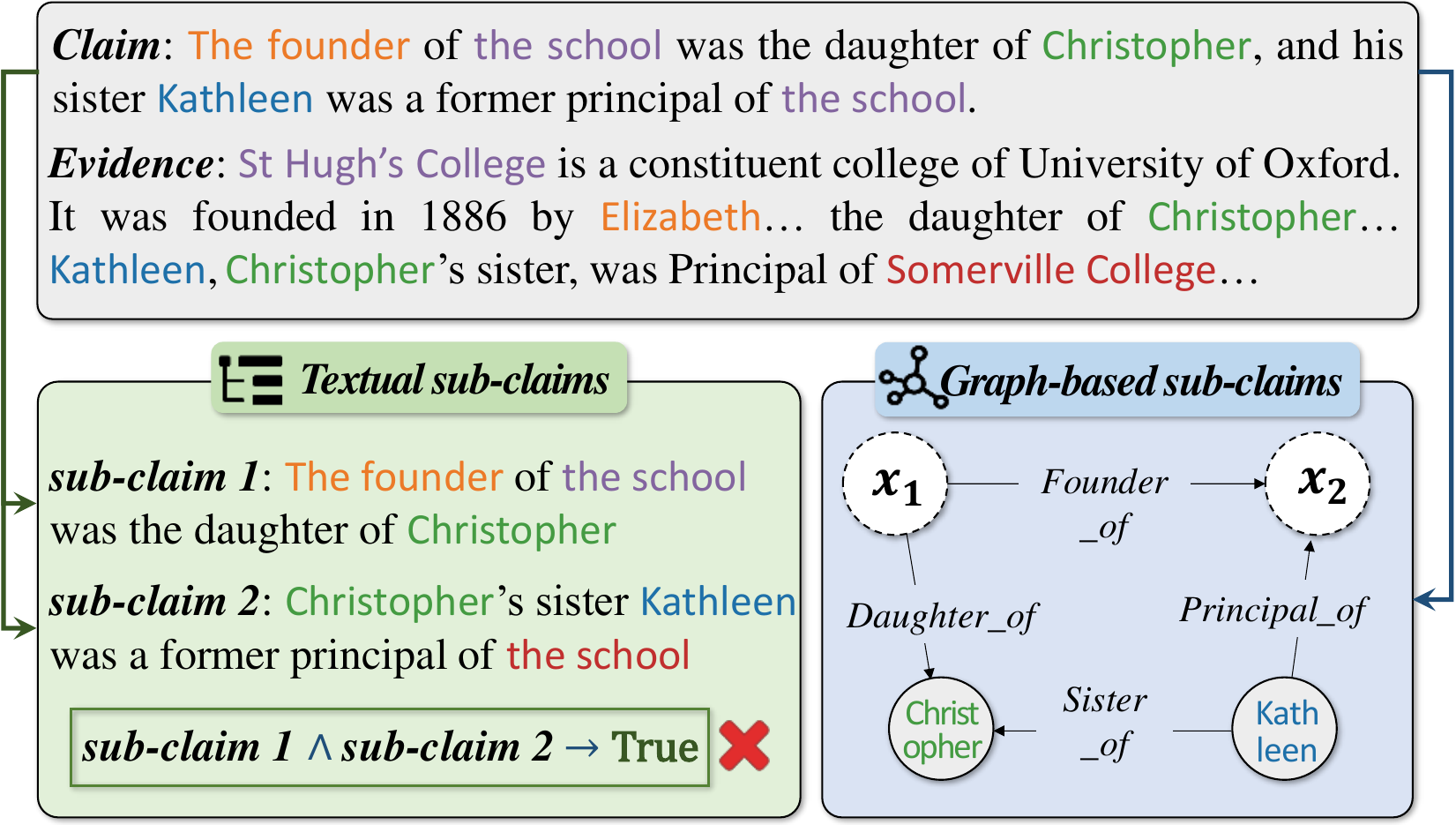}
    \caption{An illustration demonstrating the differences between existing textual sub-claims and our graph-based sub-claims. The example is from the HOVER dataset \cite{jiang2020hover}, and the decomposition results are from GPT-3.5-Turbo.}
    \label{fig:intro}
\end{figure}

Despite their impressive performance, these decomposition-based methods still face two main challenges. First, without explicit rules, existing models decompose claims into isolated textual sub-claims, each maybe containing multiple subjects, relations, or objects. This decomposition method leads to {\it insufficient decomposition} problem, introducing unnecessary complexity to the claim verification process. As in Figure \ref{fig:intro}, sub-claim 1 comprises two distinct facts: one between ``{\tt the founder}'' and ``{\tt the school}'', while the other between ``{\tt the founder}'' and ``{\tt Christopher}''.


Furthermore, there may be multiple unknown mentions within the claim text, and the same unknown mentions could be shared across different sub-claims. By focusing solely on individual sub-claims, this decomposition method increases the {\it ambiguity of mentions}, as the shared unknowns are not contextualized within a unified framework and neglects the broader context and interconnections, thereby amplifying the uncertainty associated with these unknown mentions. As in Figure \ref{fig:intro}, ``{\tt the school}'' in sub-claim 2 presents a referential ambiguity. Fact-checking models may incorrectly associate ``{\tt the school}'' with ``{\tt Somerville College}'' due to an inadequate consideration of the interdependencies between sub-claims.

To overcome the aforementioned limitations, we suggest converting the claim text into the form of graph using <subject, relation, object> triplets. The graph contains two types of entities: the known entities (the ground entities in the claim) and the unknown entities (the entities existing as references or relationships pending resolution). Using this decomposition method, each triplet can be viewed as a minimal unit that cannot be further decomposed, so the problem of \textit{insufficient decomposition} is eliminated. The graph structure also preserves all contextual information from the original claim, thus preventing \textit{ambiguity of mentions}. 

Building on such a foundation, we propose {\MyFC}, a graph-based fact-checking framework consisting of three components: (1) \textit{graph construction}, besides the claim graph, an evidence graph for each claim is constructed correspondingly, inspired by the observation that irrelevant context impairs verification accuracy from recent research \cite{guan-etal-2024-language}. (2) \textit{graph-guided planning}, prioritizes the validation of triplets and determines the logical sequence for checking each triplet. and (3) \textit{graph-guided checking}, which comprises two subtasks: \textit{graph match}, which verifies the accuracy of known entity triplets, and \textit{graph completion}, which infers incomplete entities for triplets with unknown entities.

Empirical studies employing closed-source and open-source LLMs, such as GPT-3.5-Turbo and Mistral-7B, along with widely-used fact verification datasets including HOVER \cite{jiang2020hover}, FEVEROUS \cite{aly2021feverous}, and SciFact\cite{wadden2020fact}, demonstrate the effectiveness of our proposed framework. 

To summarize, the contribution of this study is three-fold.
\begin{itemize}
    \item We introduce the graph structure to represent the complex claims, effectively addressing the insufficient decomposition problem and reducing mention ambiguity.
    \item We propose a graph-based fact-checking framework ({\MyFC}) with three components: \textit{graph construction}, \textit{graph-guided planning}, and \textit{graph-guided checking}.
    \item We validate {\MyFC}'s effectiveness and rationality through experiments on three public fact-checking datasets, achieving state-of-the-art results. 
\end{itemize}

\section{Background}
\paragraph{Fact Checking}
Given the natural-language claim $C$, fact-checking aims to find a model $M_f$ that predicts the veracity label $Y \in \{\text{True}, \text{False}\}$ of the claim $C$ based on the provided evidence $E$, which can be expressed as
\begin{align}
    M_f(C, E) \rightarrow Y.
\end{align}
Depending on how $E$ is accessed, fact-checking is usually categorized into two types: (1) Gold Evidence setting, where $E$ is directly given, and (2) Open Book setting, where $E$ needs to be retrieved from a specified knowledge source $\mathcal{K}$.

\paragraph{PLM-based Fact Checking} In the past few years, pre-trained language models have demonstrated strong performance in fact-checking. BERT-based approaches \cite{soleimani2020bert, gi2021verdict} effectively combine evidence retrieval and claim verification. Graph-based methods further enhance reasoning capabilities by modeling relationships between evidence pieces. GEAR \cite{zhou2019gear} employs evidence aggregation through graph networks, while Transformer-XH \cite{zhao2020transformer} introduces extra hop attention for multi-evidence reasoning. 

\paragraph{LLM-based Fact-Checking}
The emergence of LLMs has revolutionized fact-checking through their powerful reasoning capabilities. 
Chain-of-Thought (CoT) prompting \cite{wei2022chain} facilitates decomposing complex verification into intermediate steps, while self-consistency \cite{wang2022self} boosts accuracy via multiple reasoning paths.  
Recent work has focused on developing specialized decomposition-based prompting strategies for fact verification. HiSS \cite{zhang2023towards} proposes a hierarchical approach that decomposes claims into verifiable sub-claims, while ProgramFC \cite{pan2023fact} introduces program-guided reasoning to enhance verification accuracy. PACAR \cite{zhao2024pacar} combines planning with customized action reasoning. FOLK \cite{wang2023explainable} introduces a first-order logic-guided method for claim decomposition. 
However, these decomposition-based methods suffer from insufficient decomposition and mention ambiguity. Our work addresses these limitations by introducing a claim graph built from fine-grained sub-claims that preserves shared unknown mentions via graph structure.

\section{Method}
\begin{figure*}
    \centering
    \includegraphics[width=1\linewidth,trim={0 0 0 0},clip]{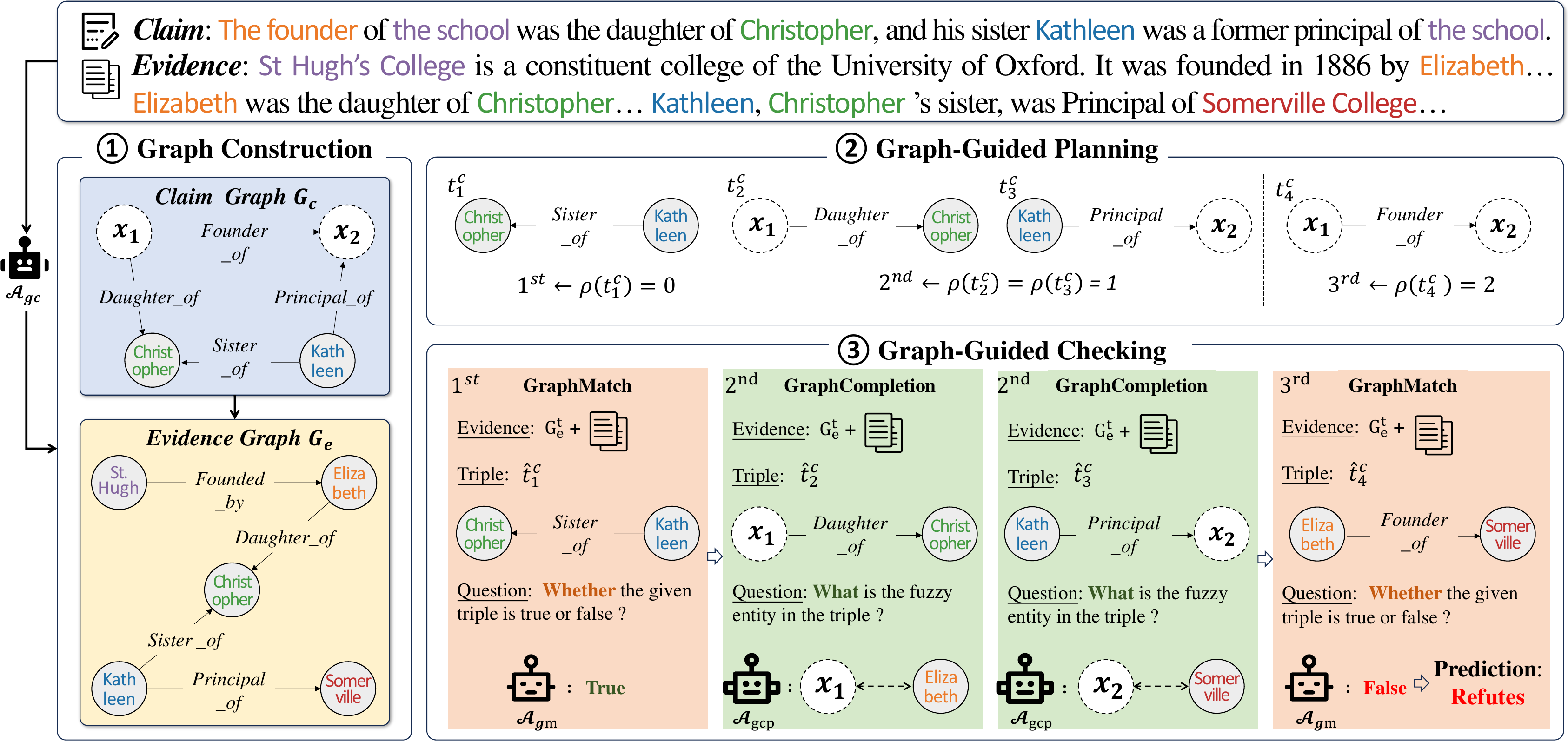}
    \caption{Overeview of {\MyFC}. The framework consists of three components: (1) \textit{graph construction} (\&\ref{sec:Graph Construction}), where claim graph and corresponding evidence graph are constructed for each claim; (2) \textit{graph-guided planning} (\&\ref{sec:Graph-Guided Planning}), which determines the verification order based on unknown entity count and plans the sequence of verification tasks; and (3) \textit{graph-guided checking} (\&\ref{sec:Graph-Guided Checking}), which executes either \textit{graph match} for verifying the known entity triples or \textit{graph completion} for inferring incomplete entity.}
    \label{fig:model}
\end{figure*}
\subsection{Fact-Checking over Graphs} \label{preliminaries}
\paragraph{Task Definition} We first give a graph-style definition for fact-checking here. Concretely, the natural language claim $C$ is converted into a directed graph $G_c = \{(s, r, o), s, o \in \mathcal{E}_c \cup \mathcal{X}_c, r \in \mathcal{R}_c\}$, where $\mathcal{E}_c$ is a known entity set, $\mathcal{X}_c$ is the unknown entity set and $\mathcal{R}_c$ is the relation set. Specifically, $\mathcal{E}_c$ contains the named or disambiguated real-world entities, which are determined when decomposing the claim. In contrast, the entity in $\mathcal{X}_c$ can not be determined according to claim only, as a claim may contain ambiguous references (e.g., pronouns, indefinite references) that cannot be immediately resolved through simple decomposition; instead, these entities require contextual understanding or additional inference steps to be determined. Similarly, the evidence set $E$ is organized into an evidence graph $G_e = \{(s, r, o), s, o \in \mathcal{E}_c, r \in \mathcal{R}_c\}$, inspired by the finding of irrelevant evidence impair verification accuracy \cite{guan-etal-2024-language}. Note that we define the nodes in $G_e$ to contain only the known entities since evidence usually comes from knowledge bases, which are written with a clear presentation of the facts, with few ambiguous references. Finally, we give the sufficient and necessary conditions for a claim to be supported by evidence.

\begin{assertion}\label{ass:graph}
A claim \( G_c \) is supported by evidence \( G_e \) if and only if there exists a one-to-one function \( \phi: \mathcal{E}_c \cup \mathcal{X}_c \to \mathcal{E}_e \) satisfies the following two conditions at the same time:
    \begin{align*}
        p_1: & \forall x \in \mathcal{E}_c, \phi(x) = x. \\
        p_2: & \forall x, y \in \mathcal{E}_c \cup \mathcal{X}_c, \forall r \in \mathcal{R}_c, \\
        & (x, r, y) \in G_c \Rightarrow (\phi(x), r, \phi(y)) \in G_e.
    \end{align*}
\end{assertion}
Based on this definition, we summarize the benefits of graph-based verification as follows: (1) is performed at the ternary level, preventing the problem of insufficient decomposition; (2) any node in the claim graph corresponds to a unique node in the evidence graph, which satisfies all the relation constraints on the node at the same time, avoiding the ambiguity of mentions.



\paragraph{{\MyFC}} While Assertion \ref{ass:graph} gives a clear idea of verification, converting it into a practical verification framework requires more effort. To validate our idea, we propose a fact-checking framework called {\bf {\MyFC}} for systematically verifying claims through graphs. As shown in Figure \ref{fig:model}, {\MyFC} consists of three primary components: {\it graph construction}, {\it graph-guided planning}, and {\it graph-guided checking}. 

\subsection{Graph Construction}\label{sec:Graph Construction}
In graph construction, we transform the claim $C$ and the corresponding evidence $E$ into the graph structures. A specialized graph construction agent $\mathcal{A}_{gc}$ is introduced to complete this complex task.


\paragraph{Claim Graph Construction}
Claim graph construction aims to transform a natural language claim $C$ to a graph $G_c$. The graph construction agent $\mathcal{A}_{gc}$ is expected to extract all triplets from claim text to complete this graph construction, which can be denoted as
\begin{align}
    G_c = \{t^c_1, \cdots, t^c_{n_c}\} = \mathcal{A}_{gc}(C)
\end{align}
where $t^c_i$ represents the $i$-th triplet extracted from the claim text and $n_c$ represents the extracted triplet number. In practice, the agent $\mathcal{A}_{gc}$ is provided with a task description, an extraction guideline and $K$ fixed in-context examples for generating triples that conform to a particular format. In particular, the unknown entities that appear in the claim are prompted to be replaced with placeholders like $x_i$ to form the triplet. Please refer to Listing \ref{lst:claim_graph} to check the prompt in detail. For example, the triplets extracted from the claim in Figure \ref{fig:model} are shown as follows:

\begin{lstlisting}[style=myStyle*]
@$t_1^c$@: <@$x_1$@, Daughter_of, Christopher>
@$t_2^c$@: <@$x_1$@, Founder_of, @$x_2$@>
@$t_3^c$@: <Kathleen, Sister_of, Christopher>
@$t_4^c$@: <Kathleen, Principal_of, @$x_2$@>
\end{lstlisting}





    
\paragraph{Evidence Graph Construction}
Evidence graph construction aims to convert the natural language evidence $E$ to a graph $G_e$. The agent $\mathcal{A}_{gc}$ in evidence graph construction process identifies entities in evidence set $E$ that are related to the known entities in $\mathcal{E}_c$ and extract the related triplets, which is denoted as
\begin{align}
    G_e = \{t^e_1, \cdots, t^e_{n_e}\} = \mathcal{A}_{gc}(E, \mathcal{E}_c)
\end{align}
where $t_i^e$ represents the $i$-th triplet extracted from the evidence text and $n_e$ represents the extracted triplet number. In practice, the agent $A_{gc}$ is provided with similar prompts and $K$ in-context examples as that for the construction of claim graphs. Please refer to Listing \ref{lst:evidence_graph} to check the prompt in detail. Even with constraints on the entity set $\mathcal{E}_c$, there are still far more triples extracted from evidence than claims, but usually, only a few of them are useful for fact verification. To improve readability, Figure \ref{fig:model} only shows the useful triples, whose output format is as follows:
\begin{lstlisting}[style=myStyle*]
@$t_1^e$@: <St Hugh's College, Founded_by, Elizabeth>
@$t_2^e$@: <Elizabeth, Daughter_of, Christopher>
@$t_3^e$@: <Kathleen, Sister_of, Christopher>
@$t_4^e$@: <Kathleen, Principal_of, Somerville>
\end{lstlisting}







\subsection{Graph-Guided Planning}\label{sec:Graph-Guided Planning}
The triplets $\{t^c_1, \cdots, t^c_{n_c}\}$ obtained from the claim graph construction are unordered. If the claim graph does not contain unknown entities (i.e., $\mathcal{X}_c=\varnothing$), it is feasible to verify the triples one by one according to any order. However, due to the presence of unknown entities, the order of validation should be planned because the triple like <$x_1$, Founder\_of, $x_2$>  will be difficult to verify before $x_1$ and $x_2$ are grounded based on known entities. In {\MyFC}, graph-guided planning is denoted as a sorting algorithm $\mathcal{S}_{gp}$, which is expressed as
\begin{align}\label{eq:planning}
    \mathcal{T} = \left[\hat{t}_i^c,\cdots,\hat{t}^c_{n_c}\right] = \mathcal{S}_{gp}(G_c)
\end{align}
where $\hat{t}_i^c$ denotes the $i$-th triplet to be verified in the subsequent step of fact-checking. Specifically, $S_{gp}$ sorts the triplets of the claim graph by the unknown entity number in the triplet. For any triplet $t=(s, p, o)$, we use $\rho(t)$ to present its priority for verification, which can be calculated by
\begin{align}
\rho(t) = \begin{cases} 
2 & \text{if } s \in \mathcal{X}_c \text{ and } o \in \mathcal{X}_c  \\
0 & \text{if } s \notin \mathcal{X}_c \text{ and } o \notin \mathcal{X}_c \\
1 & \text{otherwise}
\end{cases}
\end{align}
where the smaller value of $\rho(t)$ represents the higher priority for verification. The principle behind prioritizing in this way is summarized here: firstly, verify the triplet of two known entities only ($\rho=0$); then, ground all unknown entities from the triplet of a known entity and an unknown entity ($\rho=1$); finally, replace the unknown entities with the ground truth and verify the triplets of two unknown entities ($\rho=2$). After obtaining the priority for each triplet, we sort all triplets in the claim graph based on their priorities in descending order. The order between two triplets with the same priority will be organized randomly. Therefore, a possible ordering for the triples of the claim graph in Figure \ref{fig:model} is as follows:
\begin{lstlisting}[style=myStyle*]
@$\hat{t}_1^c$@: <Kathleen, Sister_of, Christopher>
@$\hat{t}_2^c$@: <@$x_1$@, Daughter_of, Christopher>
@$\hat{t}_3^c$@: <Kathleen, Principal_of, @$x_2$@>
@$\hat{t}_4^c$@: <@$x_1$@, Founder_of, @$x_2$@>
\end{lstlisting}

\subsection{Graph-Guided Checking}\label{sec:Graph-Guided Checking}
Based on the sorted triple list $\mathcal{T}$ in Equation \ref{eq:planning}, we conduct a fact-checking process for each triplet in the claim graph $G_c$. In general, we expect to generate a binary label $Y_{t} =\{\text{True}, \text{False}\}$ for each triplet $t$, and the fact-verification final label $Y$ with respect to the claim $C$ satisfies:
\begin{align}
    Y = Y_{\hat{t}^c_1} \wedge \cdots \wedge Y_{\hat{t}^c_{n_c}}
\end{align}
To achieve this goal, we employ two specialized components: a graph match agent $\mathcal{A}_{gm}$ for verifying the triple with two known entities and a graph completion agent $\mathcal{A}_{gcp}$ for verifying the triple with only one known entity and resolving the other unknown entity. Note that $\mathcal{A}_{gcp}$ will keep replacing unknown entities with known entities in the evidence graph. Therefore, as shown in Figure \ref{fig:model}, all triplets originally with two unknown entities will be updated to have one or two known entities, and $\mathcal{A}_{gm}$ or $\mathcal{A}_{gcp}$ can therefore solve them. We also give an algorithm-style fact-checking process in Appendix \ref{appendix:Fact-Checking over Graphs}.

\paragraph{Graph Match}\label{sec:Graph Match} When there is no unknown entity in the triplet $t$, we perform the process of graph match on it. Specifically, The graph match agent $\mathcal{A}_{gm}$ aims to verify whether the triple $t$ is supported by the evidence and output a binary label $Y_t=\{{\rm True}, {\rm False}\}$. Considering $t$ only needs to match with the related triplets in $G_e$, we filter unrelated triplets from $G_c$ to improve verification efficiency. Consequently, the triplets containing the same known entities in $t$ remain for verification, which is denoted as $G^{(t)}_e$. In addition, we empirically find that adding original evidence text $E$ as input will improve fault tolerance (See the analysis in Section \ref{Component_Analysis} for details). Therefore, the verification process of $\mathcal{A}_{gm}$ can be expressed as
\begin{align}
    Y_{t} = \mathcal{A}_{gm}(t, G^{(t)}_e, E)
\end{align}
In practice, $\mathcal{A}_{gm}$ is provided with a task description and an output-format instruction only, without the in-context examples. Taking $t^c_1$ in Figure \ref{fig:model} as an example, the simplified input and output content  are shown as follows:
\begin{lstlisting}[style=myStyle*]
# Input (@$E$@ is omitted)
@$\hat{t}_1^c$@: <Kathleen, Sister_of, Christopher>

@$G_e^{(t_1)}: \{$@
    @$t_2^e$@: <Elizabeth, Daughter_of, Christopher>
    @$t_3^e$@: <Kathleen, Sister_of, Christopher>
    @$t_4^e$@: <Kathleen, Principal_of, Somerville>
@$\}$@

# Output
@$\hat{t}_1^c \equiv t_3^e \Rightarrow Y_{t_1}=True$@ 
\end{lstlisting}

\paragraph{Graph Completion}\label{sec:Graph Completion} When there is only one unknown entity in the triplet $t$, the graph completion agent $\mathcal{A}_{gc}$ aims to ground the unknown entity $x$ in the triple and output (1) an entity $e \in \mathcal{E}_e$ to replace $x$ in all triplets of claim graph $G_c$ and (2) a binary $Y_t=\{\text{True}, \text{False}\}$. Indicating the success or failure of unknown entity completion, $Y_t$ can be simply expressed as $Y_t = (e \neq \text{None})$. Similar to $\mathcal{A}_{gm}$, we use the filtered related triplets $G_e^{(t)}$ and the original evidence text $E$ as the input. Finally, the completion process of $\mathcal{A}_{gcp}$ can be written as
\begin{align}
    e, Y_{t} = \mathcal{A}_{gcp}(t, G^{(t)}_e, E)
\end{align}
If $e$ exists, the unknown entity $x$ of the subsequent triplets should be replaced by $e$. Without loss of generality, suppose that $\mathcal{A}_{gcp}$ find the corresponding known entity $e$ for $x$ in the $i$-th triplet $\hat{t}^c_i$, for any $i<j\leq n_c$, $j$-th triplet $\hat{t}^c_j=(s, r, o) \in \mathcal{T}$ also needs to replace $x$ in it with $e$:
\begin{align}
\hat{t}^c_j =\begin{cases}
(e, r, o) & \text{if } p = x \\
(s, r, e) & \text{if } o = x \\
(s, r, o) & \text{otherwise}
\end{cases}
\end{align}
Note that while an unknown entity could theoretically exist valid mappings to multiple known entities, in practice \(\mathcal{A}_{gcp}\) typically finds a unique entity in $G_e$ due to the constraining relationships and context in $G_c$, making the determination of $e$ well-defined.
$\mathcal{A}_{gcp}$ is provided with a task description and an output-format instruction only, without the in-context examples too. Taking $t^c_2$ in Figure \ref{fig:model} as an example, the simplified input and output content  are shown as follows:
\begin{lstlisting}[style=myStyle*]
# Input (@$E$@ is omitted)
@$\hat{t}_2^c$@: <@$x_1$@, Daughter_of, Christopher>

@$G_e^{(t_2)}=\{$@
    @$t_2^e$@: <Elizabeth, Daughter_of, Christopher>
    @$t_3^e$@: <Kathleen, Sister_of, Christopher>
@$\}$@

# Output
@$e \rightarrow Elizabeth \Rightarrow Y_{t_2} = (x_1 != {\rm None}) = {\rm True}$@

# Update other triplets
@$\hat{t}_4^c$@: <@$x_1$@, Founder_of, @$x_2$@>@$\leftarrow$@<Elizabeth, Founder_of, @$x_2$@>
\end{lstlisting}

\paragraph{Early Stop in Checking} 
The claim requires all triples to be verified as True for support. In practice, if any triple validates as False, we terminate the process early and set the final label $Y$ to False.

\section{Experiment}
\subsection{Experiment Setup}
\paragraph{Datasets} 
We evaluate {\MyFC} on three widely used fact-checking datasets. \textbf{HOVER} \cite{jiang2020hover} comprises claims necessitating multi-hop reasoning across Wikipedia articles for validation. We employ its validation set, which is categorized into three subsets according to reasoning complexity: two-hop, three-hop, and four-hop. For \textbf{FEVEROUS} \cite{aly2021feverous}, we use its validation set and select claims that only require sentence evidence following~\cite{pan2023fact}. For \textbf{SciFact} \cite{wadden2020fact}, we employ its validation set, where claims with complete evidence that either support or refute the claim are selected. 
These datasets encompass various domains and complexity levels, offering a thorough benchmark for automated fact-checking systems.
\paragraph{Baselines}
We evaluate {\MyFC} against ten strong baselines in two categories. The first category comprises \textbf{pretrained/fine-tuned models}, including \underline{BERT-FC} \cite{soleimani2020bert}, which uses BERT-large for binary classification; \underline{LisT5} \cite{jiang2021exploring}, leveraging T5 with listwise concatenation; \underline{RoBERTa-NLI} \cite{nie2019combining} and \underline{DeBERTaV3-NLI} \cite{he2021debertav3}, both fine-tuned on multiple NLI datasets; and \underline{MULTIVERS} \cite{wadden2021multivers}, which employs LongFormer \cite{beltagy2020longformer} for processing long evidence sequences. The second category consists of \textbf{in-context learning models}, including \underline{Codex} \cite{chen2021evaluating} and \underline{FLAN-T5} \cite{chung2022scaling} with 20-shot learning, \underline{ProgramFC} \cite{pan2023fact} combines Codex and FLAN-T5 for generating reasoning programs. We also implement two prompt-based approaches: \underline{Direct} (zero-shot) and \underline{Decomposition} (10-shot), which break claims into multiple textual sub-claims, using GPT-3.5-Turbo. Although other notable methods like PACAR \cite{zhao2024pacar} exist, they are excluded due to differences in retrieval techniques and the lack of open-source code, ensuring a fair comparison. Details about the baselines are provided in Appendix \ref{appendix:baselines}

\paragraph{Implementation Details}
We use \texttt{gpt-3.5-turbo-0125} as our main language model, with \texttt{Mistral-7B-Instruct-v0.3}\footnote{https://huggingface.co/mistralai/Mistral-7B-Instruct-v0.3} for ablation study \ref{Backbone_Analysis}. Experiments run on a Tesla V100 SXM2 32GB GPU with Intel(R) Xeon(R) Silver 4214 CPU.
For Graph Construction, we implement 10-shot learning ($K$=10), more challenging than the 20 examples used by baselines (Codex, FLAN-T5, ProgramFC). Examples are randomly sampled from the training set. Graph completion and matching operate in a zero-shot mode without demonstrations.
In the Gold Evidence setting (gold), we use provided gold evidence. In the Open Book setting (open), we follow ProgramFC \cite{pan2023fact}, using BM25 \cite{robertson2009probabilistic} retriever via Pyserini \cite{lin2021pyserini} for all methods. The top-5 retrieved paragraphs serve as evidence.



\paragraph{Evaluation Metric}
Following \cite{pan2023fact} and \cite{wang2023explainable}, we employ the macro-F1 score as the primary metric, which provides a balanced evaluation across all categories.

\subsection{Main Results}\label{sec:main_results}
\begin{table*}[t]

\centering
\renewcommand{\arraystretch}{0.9}
\resizebox{\textwidth}{!}{
\begin{tabular}{l|c c c c c c|c c|c c}
\hline
\multirow{2}{*}{Model} & \multicolumn{2}{c}{HOVER(2-hop)} & \multicolumn{2}{c}{HOVER(3-hop)}& \multicolumn{2}{c|}{HOVER(4-hop)} & \multicolumn{2}{c|}{FEVEROUS} & \multicolumn{2}{c}{SciFact} \\


& gold & open & gold & open & gold & open & gold & open & gold & open \\
\hline
\rowcolor{gray!10}
\multicolumn{11}{c}{\textit{Pretrained/Fine-tuned}} \\
BERT-FC & 53.40 & 50.68 & 50.90 & 49.86 & 50.86 & 48.57 & 74.71 & 51.60 & - & - \\
LisT5 & 56.15 & 52.56 & 53.76 & 51.89 & 51.67 & 50.46 & 77.88 & 54.15 & - & - \\
RoBERTa-NLI & 74.62 & 63.62 & 62.23 & 53.99 & 57.98 & 52.40 & 88.28 & 57.80 & - & - \\
DeBERTaV3-NLI & \underline{77.22} & 68.72 & 65.98 & 60.76 & 60.49 & 56.00 & \underline{91.98} & 58.81 & - & - \\
MULTIVERS & 68.86 & 60.17 & 59.87 & 52.55 & 55.67 & 51.86 & 86.03 & 56.61 & 77.24 & 54.17 \\
\rowcolor{gray!10}
\multicolumn{11}{c}{\textit{In-context learning}} \\
Codex & 70.63 & 65.07 & 66.46 & 56.63 & 63.49 & 57.27 & 89.77 & 62.58 & - & - \\
FLAN-T5 & 73.69 & 69.02 & 65.66 & 60.23 & 58.08 & 55.42 & 90.81 & 63.73 & - & - \\
ProgramFC & 74.10 & \underline{69.36} & 66.13 & \underline{60.63} & \underline{65.69} & \underline{59.16} & 91.77 & \underline{67.80} & \underline{84.90} & 70.63 \\

Direct & 73.53 & 68.85 & 65.09 & 53.77 & 59.03 & 46.93 & 88.44 & 66.41 & 73.30 & 60.54 \\
Decompostion & 72.67 & 65.40 & \underline{66.29} & 54.65 & 60.88 & 54.41 & 85.79 & 59.38 & 81.64 & \underline{72.92} \\
\hline
\rowcolor{gray!50}
\textbf{{\MyFC}} & \textbf{77.85} & \textbf{73.44} & \textbf{70.35} & \textbf{66.54} & \textbf{71.42} & \textbf{67.47} & \textbf{93.75} & \textbf{72.88} & \textbf{87.37} & \textbf{80.63} \\
\hline
\end{tabular}}
\caption{Comparison of main results (macro-F1 in \%) across the HOVER, FEVEROUS, and SciFact datasets in both open-book and gold evidence settings. The best results are in \textbf{bold}, and the second-best results are \underline{underlined}.}
\label{tab:main_results}
\end{table*}
\begin{figure*}[!t]
    \centering
    \subfloat[Component analysis of {\MyFC} on HOVER in terms of macro-F1.]
    {
        \includegraphics[width=0.48\linewidth]{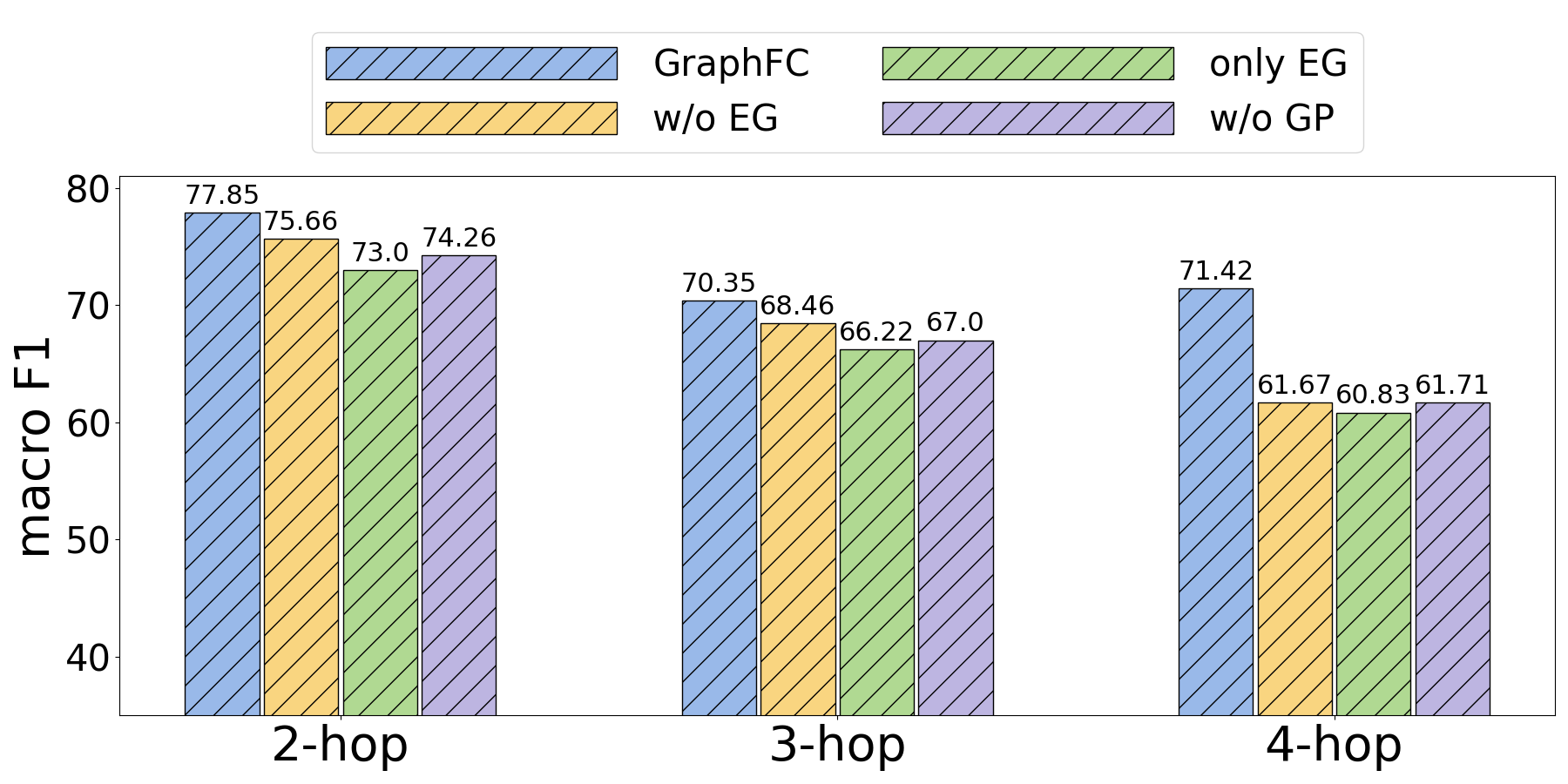}
        \label{fig:abl_1}
    }
    \hfill
    \subfloat[Backbone analysis  of {\MyFC} (GPT-3.5 vs Mistral-7B) on HOVER in terms of macro-F1.]
    {
        \includegraphics[width=0.48\linewidth]{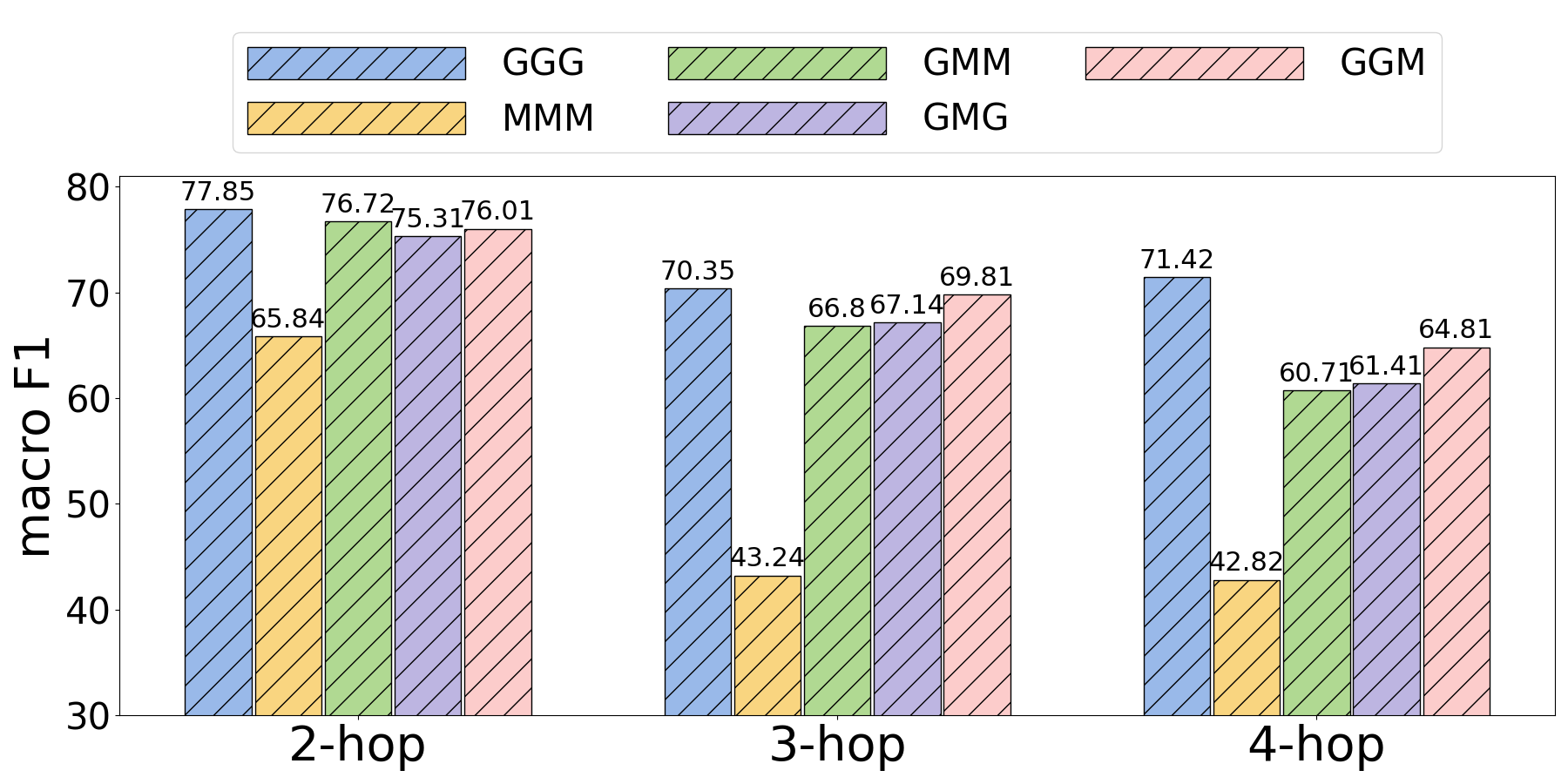}
        \label{fig:abl_2}
    }
    \caption{Ablation studies of {\MyFC}.}
    \label{fig:ablation}
\end{figure*}

{\MyFC} significantly outperforms state-of-the-art baselines, as shown in Table~\ref{tab:main_results}, highlighting three key findings that confirm its effectiveness.

\noindent {\textbf{{\MyFC} consistently outperforms existing methods across all  settings.}} On HOVER, we achieve accuracy improvements of 3.75-5.73\% in gold settings and 4.08-8.31\% in open-book settings across 2-4 hop claims compared to ProgramFC. For FEVEROUS, our method outperforming ProgramFC by 1.98\% and 5.08\% in gold and open settings, respectively. On SciFact, we obtain 2.47\% and 7.71\% gains in gold and open settings, respectively.

\noindent {\textbf{{\MyFC} shows increasing performance gains with claim complexity.}} Results on HOVER shows that performance improvements increase with reasoning complexity, from 4.08\% for 2-hop to 8.31\% for 4-hop claims in open setting. This validates the effectiveness of our graph-based approach in handling complex claims.

\noindent {\textbf{{\MyFC} maintains robust performance in open setting.}} 
The performance gap between open and gold setting is smaller with {\MyFC}, showing a 3.95\% gap for 4-hop claims on HOVER compared to ProgramFC's 6.53\%. This improvement stems from our evidence graph construction that effectively filters and organizes relevant information to enhance reasoning.


\subsection{Ablation Study}\label{sec:ablation_study}

\subsubsection{Component Analysis}\label{Component_Analysis}
To evaluate the effectiveness of key components in {\MyFC}, we conduct ablation experiments by removing or modifying specific components. As shown in Figure \ref{fig:abl_1}, the variant \textit{w/o EG} eliminates the \underline{E}vidence \underline{G}raph Construction component, directly performing fact-checking on raw evidence texts. \textit{only EG} exclusively utilizes the constructed evidence graph without referencing the original text. \textit{w/o GP} removes the \underline{G}raph-Guided \underline{P}lanning component, leading to independent parallel verification of each triple in the claim graph. 

\noindent {\textbf{Effectiveness of evidence Graph construction.}} Removing the evidence graph (\textit{w/o EG}) causes performance drops at all hop levels on HOVER, especially a significant decrease at 4 hops claims, which contain richer knowledge and longer text, requiring extensive evidence for verification. This demonstrates the advantage of evidence graphs in constructing structured evidence that effectively filters out irrelevant context. On the other hand, using only the evidence graph (\textit{only EG}) leads to performance loss at all levels, indicating that relying solely on the graph can result in the loss of valuable information from the evidence text.

\noindent {\textbf{Effectiveness of graph-guided planning.}} Removing graph-guided planning (\textit{w/o GP}) significantly decreases overall performance, especially in 4-hop claims on HOVER which require multi-step reasoning, highlighting its crucial role in structuring multi-step reasoning and ensuring that unknown entities are properly identified in complex verification tasks.


\begin{table*}[!hbt]
    \centering
    \includegraphics[width=\linewidth]{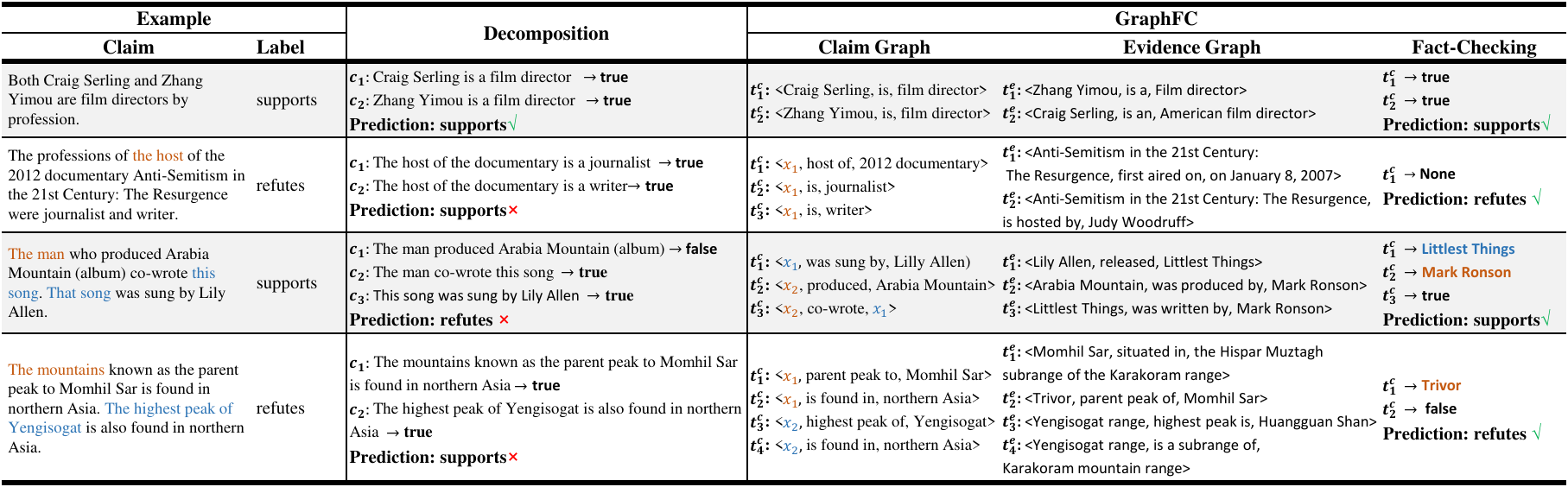}
    \caption{Case study of representative examples from HOVER comparing Decomposition and {\MyFC}. }
    \label{tab:case_study}
    \vspace{-5mm}
\end{table*}
\subsubsection{Backbone Analysis} \label{Backbone_Analysis}
To investigate the impact of different language models on model performance, we conducted ablation experiments by replacing GPT-3.5 Turbo with Mistral-7B across three components: graph construction, matching, and completion agents. The baseline configuration using GPT-3.5 for all components is denoted as \textit{GGG}, while other configurations include \textit{MMM} (all Mistral-7B), \textit{GMM}, \textit{GMG}, and \textit{GGM} (combinations of both models).

\noindent {\textbf{Graph construction component critically depends on advanced LLM capabilities.}} The full Mistral-7B configuration (\textit{MMM}) shows significant performance degradation, with 2-hop accuracy dropping from 77.85\% to 65.84\%. Graph construction requires strong reasoning abilities to decompose complex claims into fine-grained triples.

\begin{figure}[h]
    \centering
    \includegraphics[width=1\linewidth]{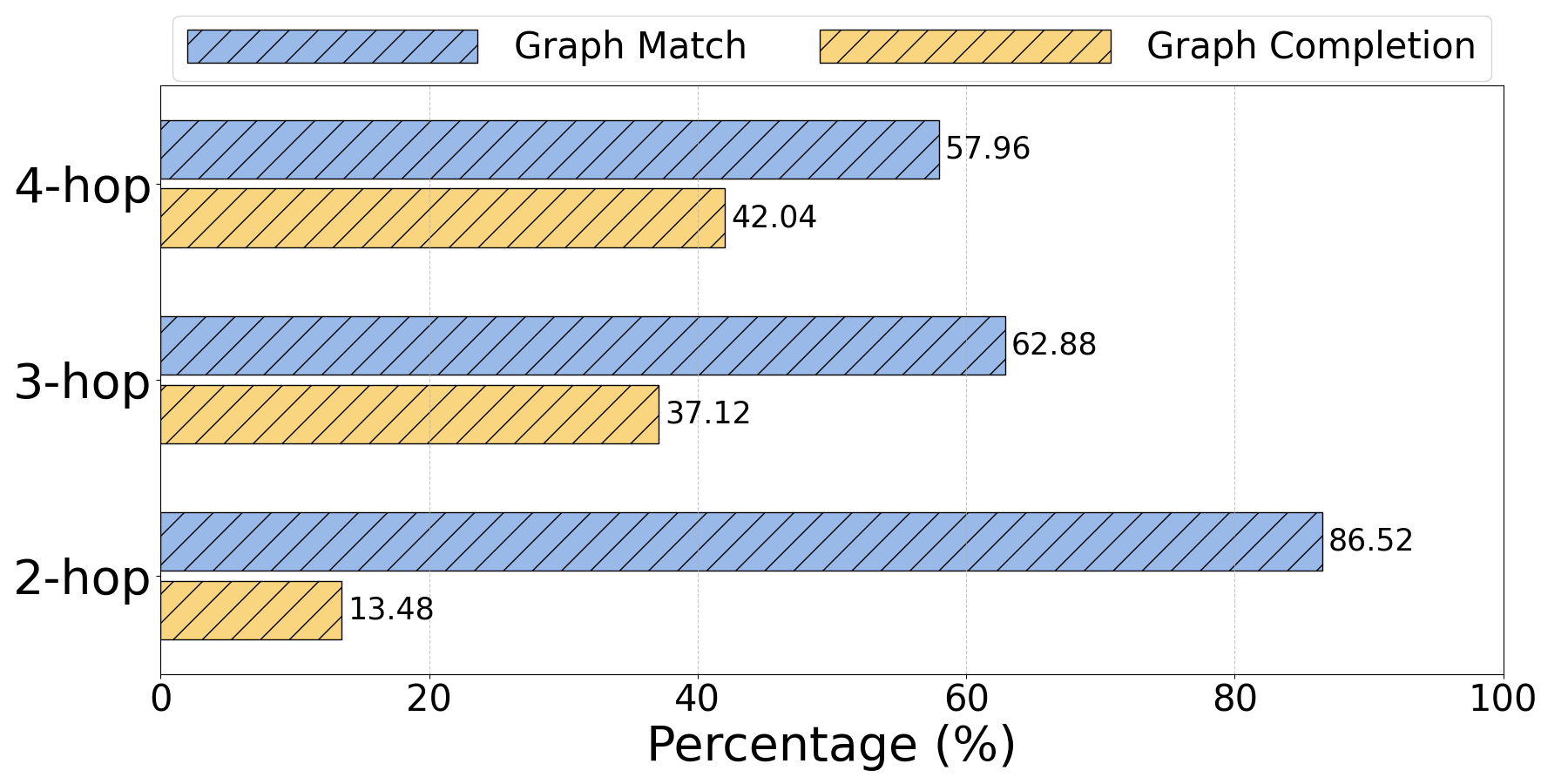}
    \caption{Distribution of task proportions for graph match and graph completion on HOVER}
    \label{fig:dist}
\end{figure}

\noindent {\textbf{Graph match component shows higher model sensitivity than Completion.}} The graph matching configuration (\textit{GMG}'s) leads to a notable drop in 2-hop accuracy to 75.31\%, while the completion component (\textit{GGM}'s) remains relatively robust with accuracy only decreasing to 76.01\%. This aligns with the distribution of reasoning tasks on HOVER, as shown in Figure \ref{fig:dist}, where matching operations dominate verification tasks across all hops. These findings suggest that while graph construction requires powerful models, other components can utilize lighter models when prioritizing computational efficiency over maximum accuracy.

\subsection{Case Study}\label{sec:case_study}
To demonstrate {\MyFC}'s ability to address the limitations of existing decomposition-based methods, we compare it with the baseline Decomposition method using four examples of varying complexity in Table~\ref{tab:case_study}. For clarity, only essential triples from the evidence graph that directly support or refute the claim are shown.
The first example highlights the ability of both methods to handle simple claims with clear entity relationships. The third example exposes a major limitation of the Decomposition method: when verifying the sub-claim about ``Arabia Mountain's producer," it fails to maintain contextual relationships between decomposed sub-claims, resulting in ambiguity around "the man." In contrast, {\MyFC} preserves these connections through its graph structure, explicitly modeling entities and their relationships as triples. Additionally, graph-guided planning optimizes the verification order, positioning triples with unknown entities at both ends for later processing. It demonstrates that {\MyFC} improves accuracy by preserving contextual relationships and ensuring an optimal verification sequence.

\section{Conclusion}
In this paper, we proposed {\MyFC}, a graph-based fact-checking framework that effectively addresses two critical challenges in LLM-driven decomposition-based method: insufficient decomposition and mention ambiguity. By converting claims into graph structures, our approach preserves contextual relationships while enabling fine-grained verification through minimal triplet units. Experimental results demonstrate GraphFC's effectiveness across three benchmarks, achieving state-of-the-art performance with significant improvements in complex reasoning scenarios. Ablation studies highlight the crucial role of graph construction and graph-guided planning in handling complex claims. These findings establish GraphFC as an effective approach for enhancing automated fact-checking systems.

\section*{Limitations}
While {\MyFC} demonstrates substantial improvements over existing approaches, it exhibits limitations in two aspects. 

\begin{itemize}
    \item  Our approach achieves relatively modest gains for simple claims verification. As shown in Table~\ref{tab:main_results}, the improvement margin on 2-hop claims in the HOVER dataset (0.63\% in gold setting) is notably smaller compared to 3-hop and 4-hop claims. This limitation is inherent in our method's design: while {\MyFC} excels at handling complex multi-hop reasoning, for simple claims where conventional methods already perform well, the sophisticated graph construction and matching procedures may introduce unnecessary complexity.
    \item The graph construction component of our method relies on the reasoning capabilities of LLMs. As demonstrated in Figure~\ref{fig:abl_2}, replacing GPT-3.5 with Mistral-7B in the graph construction phase leads to a significant decline in performance. This suggests that the accuracy of decomposing claims into fine-grained triple representations benefits from the advanced reasoning capabilities of more powerful language models. Future work will focus on developing lighter-weight graph construction techniques and exploring more efficient verification strategies for simple claims while maintaining the current effectiveness on complex reasoning tasks.
\end{itemize}

\bibliography{acl2023}

\begin{thebibliography}{30}
\expandafter\ifx\csname natexlab\endcsname\relax\def\natexlab#1{#1}\fi

\bibitem[{Aly et~al.(2021)Aly, Guo, Schlichtkrull, Thorne, Vlachos, Christodoulopoulos, Cocarascu, and Mittal}]{aly2021feverous}
Rami Aly, Zhijiang Guo, Michael Schlichtkrull, James Thorne, Andreas Vlachos, Christos Christodoulopoulos, Oana Cocarascu, and Arpit Mittal. 2021.
\newblock Feverous: Fact extraction and verification over unstructured and structured information.
\newblock \emph{arXiv preprint arXiv:2106.05707}.

\bibitem[{Beltagy et~al.(2020)Beltagy, Peters, and Cohan}]{beltagy2020longformer}
Iz~Beltagy, Matthew~E Peters, and Arman Cohan. 2020.
\newblock Longformer: The long-document transformer.
\newblock \emph{arXiv preprint arXiv:2004.05150}.

\bibitem[{Bowman et~al.(2015)Bowman, Angeli, Potts, and Manning}]{DBLP:conf/emnlp/BowmanAPM15}
Samuel~R. Bowman, Gabor Angeli, Christopher Potts, and Christopher~D. Manning. 2015.
\newblock \href {https://doi.org/10.18653/v1/D15-1075} {A large annotated corpus for learning natural language inference}.
\newblock In \emph{Proceedings of the 2015 Conference on Empirical Methods in Natural Language Processing (EMNLP)}, pages 632--642, Lisbon, Portugal.

\bibitem[{Chen et~al.(2021)Chen, Tworek, Jun, Yuan, Pinto, Kaplan, Edwards, Burda, Joseph, Brockman et~al.}]{chen2021evaluating}
Mark Chen, Jerry Tworek, Heewoo Jun, Qiming Yuan, Henrique Ponde De~Oliveira Pinto, Jared Kaplan, Harri Edwards, Yuri Burda, Nicholas Joseph, Greg Brockman, et~al. 2021.
\newblock Evaluating large language models trained on code.
\newblock \emph{arXiv preprint arXiv:2107.03374}.

\bibitem[{Chung et~al.(2022)Chung, Hou, Longpre, Zoph, Tay, Fedus, Li, Wang, Dehghani, Brahma et~al.}]{chung2022scaling}
Hyung~Won Chung, Le~Hou, Shayne Longpre, Barret Zoph, Yi~Tay, William Fedus, Yunxuan Li, Xuezhi Wang, Mostafa Dehghani, Siddhartha Brahma, et~al. 2022.
\newblock Scaling instruction-finetuned language models.
\newblock \emph{arXiv preprint arXiv:2210.11416}.

\bibitem[{Gi et~al.(2021)Gi, Fang, and Tsai}]{gi2021verdict}
In-Zu Gi, Ting-Yu Fang, and Richard Tzong-Han Tsai. 2021.
\newblock Verdict inference with claim and retrieved elements using roberta.
\newblock In \emph{Proceedings of the Fourth Workshop on Fact Extraction and VERification (FEVER)}, pages 60--65.

\bibitem[{Guan et~al.(2024)Guan, Dodge, Wadden, Huang, and Peng}]{guan-etal-2024-language}
Jian Guan, Jesse Dodge, David Wadden, Minlie Huang, and Hao Peng. 2024.
\newblock \href {https://doi.org/10.18653/v1/2024.naacl-long.62} {Language models hallucinate, but may excel at fact verification}.
\newblock In \emph{Proceedings of the 2024 Conference of the North American Chapter of the Association for Computational Linguistics: Human Language Technologies (Volume 1: Long Papers)}, pages 1090--1111, Mexico City, Mexico. Association for Computational Linguistics.

\bibitem[{He et~al.(2021)He, Gao, and Chen}]{he2021debertav3}
Pengcheng He, Jianfeng Gao, and Weizhu Chen. 2021.
\newblock Debertav3: Improving deberta using electra-style pre-training with gradient-disentangled embedding sharing.
\newblock \emph{arXiv preprint arXiv:2111.09543}.

\bibitem[{Jiang et~al.(2021)Jiang, Pradeep, and Lin}]{jiang2021exploring}
Kelvin Jiang, Ronak Pradeep, and Jimmy Lin. 2021.
\newblock Exploring listwise evidence reasoning with t5 for fact verification.
\newblock In \emph{Proceedings of the 59th Annual Meeting of the Association for Computational Linguistics and the 11th International Joint Conference on Natural Language Processing (Volume 2: Short Papers)}, pages 402--410.

\bibitem[{Jiang et~al.(2020)Jiang, Bordia, Zhong, Dognin, Singh, and Bansal}]{jiang2020hover}
Yichen Jiang, Shikha Bordia, Zheng Zhong, Charles Dognin, Maneesh Singh, and Mohit Bansal. 2020.
\newblock Hover: A dataset for many-hop fact extraction and claim verification.
\newblock \emph{arXiv preprint arXiv:2011.03088}.

\bibitem[{Lin et~al.(2021)Lin, Ma, Lin, Yang, Pradeep, and Nogueira}]{lin2021pyserini}
Jimmy Lin, Xueguang Ma, Sheng-Chieh Lin, Jheng-Hong Yang, Ronak Pradeep, and Rodrigo Nogueira. 2021.
\newblock Pyserini: A python toolkit for reproducible information retrieval research with sparse and dense representations.
\newblock In \emph{Proceedings of the 44th International ACM SIGIR Conference on Research and Development in Information Retrieval}, pages 2356--2362.

\bibitem[{Liu et~al.(2022)Liu, Swayamdipta, Smith, and Choi}]{DBLP:journals/corr/abs-2201-05955}
Alisa Liu, Swabha Swayamdipta, Noah~A. Smith, and Yejin Choi. 2022.
\newblock \href {https://aclanthology.org/2022.findings-emnlp.508} {{WANLI}: Worker and {AI} collaboration for natural language inference dataset creation}.
\newblock In \emph{Findings of the Association for Computational Linguistics: EMNLP 2022}, pages 6826--6847, Abu Dhabi, United Arab Emirates.

\bibitem[{Nie et~al.(2019{\natexlab{a}})Nie, Chen, and Bansal}]{nie2019combining}
Yixin Nie, Haonan Chen, and Mohit Bansal. 2019{\natexlab{a}}.
\newblock Combining fact extraction and verification with neural semantic matching networks.
\newblock In \emph{Proceedings of the AAAI conference on artificial intelligence}, pages 6859--6866.

\bibitem[{Nie et~al.(2019{\natexlab{b}})Nie, Chen, and Bansal}]{DBLP:conf/aaai/NieCB19}
Yixin Nie, Haonan Chen, and Mohit Bansal. 2019{\natexlab{b}}.
\newblock \href {https://doi.org/10.1609/aaai.v33i01.33016859} {Combining fact extraction and verification with neural semantic matching networks}.
\newblock In \emph{Proceedings of the 33rd {AAAI} Conference on Artificial Intelligence (AAAI)}, pages 6859--6866, Honolulu, Hawaii, USA.

\bibitem[{Nie et~al.(2020)Nie, Williams, Dinan, Bansal, Weston, and Kiela}]{DBLP:conf/acl/NieWDBWK20}
Yixin Nie, Adina Williams, Emily Dinan, Mohit Bansal, Jason Weston, and Douwe Kiela. 2020.
\newblock \href {https://doi.org/10.18653/v1/2020.acl-main.441} {Adversarial {NLI}: A new benchmark for natural language understanding}.
\newblock In \emph{Proceedings of the 58th Annual Meeting of the Association for Computational Linguistics (ACL)}, pages 4885--4901, Online.

\bibitem[{Pan et~al.(2023)Pan, Wu, Lu, Luu, Wang, Kan, and Nakov}]{pan2023fact}
Liangming Pan, Xiaobao Wu, Xinyuan Lu, Anh~Tuan Luu, William~Yang Wang, Min-Yen Kan, and Preslav Nakov. 2023.
\newblock Fact-checking complex claims with program-guided reasoning.
\newblock \emph{arXiv preprint arXiv:2305.12744}.

\bibitem[{Parrish et~al.(2021)Parrish, Huang, Agha, Lee, Nangia, Warstadt, Aggarwal, Allaway, Linzen, and Bowman}]{DBLP:conf/emnlp/ParrishHALNWAAL21}
Alicia Parrish, William Huang, Omar Agha, Soo-Hwan Lee, Nikita Nangia, Alexia Warstadt, Karmanya Aggarwal, Emily Allaway, Tal Linzen, and Samuel~R. Bowman. 2021.
\newblock \href {https://doi.org/10.18653/v1/2021.findings-emnlp.421} {Does putting a linguist in the loop improve {NLU} data collection?}
\newblock In \emph{Findings of the Association for Computational Linguistics: EMNLP 2021}, pages 4886--4901, Punta Cana, Dominican Republic.

\bibitem[{Robertson et~al.(2009)Robertson, Zaragoza et~al.}]{robertson2009probabilistic}
Stephen Robertson, Hugo Zaragoza, et~al. 2009.
\newblock The probabilistic relevance framework: Bm25 and beyond.
\newblock \emph{Foundations and Trends{\textregistered} in Information Retrieval}, 3(4):333--389.

\bibitem[{Soleimani et~al.(2020)Soleimani, Monz, and Worring}]{soleimani2020bert}
Amir Soleimani, Christof Monz, and Marcel Worring. 2020.
\newblock Bert for evidence retrieval and claim verification.
\newblock In \emph{Advances in Information Retrieval: 42nd European Conference on IR Research, ECIR 2020, Lisbon, Portugal, April 14--17, 2020, Proceedings, Part II 42}, pages 359--366. Springer.

\bibitem[{Thorne et~al.(2018)Thorne, Vlachos, Christodoulopoulos, and Mittal}]{thorne2018fever}
James Thorne, Andreas Vlachos, Christos Christodoulopoulos, and Arpit Mittal. 2018.
\newblock Fever: a large-scale dataset for fact extraction and verification.
\newblock \emph{arXiv preprint arXiv:1803.05355}.

\bibitem[{Wadden et~al.(2020)Wadden, Lin, Lo, Wang, van Zuylen, Cohan, and Hajishirzi}]{wadden2020fact}
David Wadden, Shanchuan Lin, Kyle Lo, Lucy~Lu Wang, Madeleine van Zuylen, Arman Cohan, and Hannaneh Hajishirzi. 2020.
\newblock Fact or fiction: Verifying scientific claims.
\newblock \emph{arXiv preprint arXiv:2004.14974}.

\bibitem[{Wadden et~al.(2021)Wadden, Lo, Wang, Cohan, Beltagy, and Hajishirzi}]{wadden2021multivers}
David Wadden, Kyle Lo, Lucy~Lu Wang, Arman Cohan, Iz~Beltagy, and Hannaneh Hajishirzi. 2021.
\newblock Multivers: Improving scientific claim verification with weak supervision and full-document context.
\newblock \emph{arXiv preprint arXiv:2112.01640}.

\bibitem[{Wang and Shu(2023)}]{wang2023explainable}
Haoran Wang and Kai Shu. 2023.
\newblock Explainable claim verification via knowledge-grounded reasoning with large language models.
\newblock \emph{arXiv preprint arXiv:2310.05253}.

\bibitem[{Wang et~al.(2022)Wang, Wei, Schuurmans, Le, Chi, Narang, Chowdhery, and Zhou}]{wang2022self}
Xuezhi Wang, Jason Wei, Dale Schuurmans, Quoc Le, Ed~Chi, Sharan Narang, Aakanksha Chowdhery, and Denny Zhou. 2022.
\newblock Self-consistency improves chain of thought reasoning in language models.
\newblock \emph{arXiv preprint arXiv:2203.11171}.

\bibitem[{Wei et~al.(2022)Wei, Wang, Schuurmans, Bosma, Xia, Chi, Le, Zhou et~al.}]{wei2022chain}
Jason Wei, Xuezhi Wang, Dale Schuurmans, Maarten Bosma, Fei Xia, Ed~Chi, Quoc~V Le, Denny Zhou, et~al. 2022.
\newblock Chain-of-thought prompting elicits reasoning in large language models.
\newblock \emph{Advances in neural information processing systems}, 35:24824--24837.

\bibitem[{Williams et~al.(2018)Williams, Nangia, and Bowman}]{DBLP:conf/naacl/WilliamsNB18}
Adina Williams, Nikita Nangia, and Samuel Bowman. 2018.
\newblock \href {https://doi.org/10.18653/v1/N18-1101} {A broad-coverage challenge corpus for sentence understanding through inference}.
\newblock In \emph{Proceedings of the 2018 Conference of the North {A}merican Chapter of the Association for Computational Linguistics: Human Language Technologies (NAACL-HLT)}, pages 1112--1122, New Orleans, Louisiana, USA.

\bibitem[{Zhang and Gao(2023)}]{zhang2023towards}
Xuan Zhang and Wei Gao. 2023.
\newblock Towards llm-based fact verification on news claims with a hierarchical step-by-step prompting method.
\newblock \emph{arXiv preprint arXiv:2310.00305}.

\bibitem[{Zhao et~al.(2020)Zhao, Xiong, Rosset, Song, Bennett, and Tiwary}]{zhao2020transformer}
Chen Zhao, Chenyan Xiong, Corby Rosset, Xia Song, Paul Bennett, and Saurabh Tiwary. 2020.
\newblock Transformer-xh: Multi-evidence reasoning with extra hop attention.
\newblock In \emph{International Conference on Learning Representations}.

\bibitem[{Zhao et~al.(2024)Zhao, Wang, Wang, Cheng, Zhang, and Wong}]{zhao2024pacar}
Xiaoyan Zhao, Lingzhi Wang, Zhanghao Wang, Hong Cheng, Rui Zhang, and Kam-Fai Wong. 2024.
\newblock Pacar: Automated fact-checking with planning and customized action reasoning using large language models.
\newblock In \emph{Proceedings of the 2024 Joint International Conference on Computational Linguistics, Language Resources and Evaluation (LREC-COLING 2024)}, pages 12564--12573.

\bibitem[{Zhou et~al.(2019)Zhou, Han, Yang, Liu, Wang, Li, and Sun}]{zhou2019gear}
Jie Zhou, Xu~Han, Cheng Yang, Zhiyuan Liu, Lifeng Wang, Changcheng Li, and Maosong Sun. 2019.
\newblock Gear: Graph-based evidence aggregating and reasoning for fact verification.
\newblock \emph{arXiv preprint arXiv:1908.01843}.

\end{thebibliography}
\bibliographystyle{acl_natbib}

\appendix
\section*{Appendix}
\section{Graph-Guided Fact-Checking}\label{appendix:Fact-Checking over Graphs}
Here we give an algorithm-style graph-guided fact-checking process in algorithm \ref{alg:1}. Based on the sorted triple list $\mathcal{T}$ in Equation \ref{eq:planning}, we conduct a fact-checking process for each triplet in the claim graph $G_c$. 
\begin{enumerate}
    \item If the triple contains only grounded entities, we apply a graph match agent to verify its validity against \( G_e^{(t)} \). Under this situation, if the validation process returns a {\em false}, then the fact-checking function returns a {\em false}.
    \item Otherwise, if the triple contains an unknown entity, we apply a graph completion agent to find the correct entity and ground it. If grounding is successful, we update both the claim graph $G_c$ and evidence graph $G_e$. If the function fails to find grounding results, the fact-checking function returns {\em false}.
    \item  If no {\em false} returns during the fact-checking process, then the fact-checking function returns a {\em true}.
\end{enumerate}

\begin{algorithm}[ht]
\caption{Graph-Guided Fact-Checking}
\label{alg:1}
\begin{algorithmic}
\REQUIRE claim graph \( G_c \) , evidence graph \( G_e \), evidence set \(E\), agents $\mathcal{A}_{gc}$, $\mathcal{A}_{gm}$, $\mathcal{A}_{gcp}$, and sorted triples $\mathcal{T}$
\ENSURE fact-checking result $Y$ (True or False)
\STATE \( Y \leftarrow \text{True} \)
\FOR{each \( t \) in \( \mathcal{T} \)}
    \STATE \( (s, p, o) \leftarrow t \)
    \IF{  $s \notin \mathcal{X}_c \text{ and } o\notin \mathcal{X}_c$}
            \STATE \(Y_{t} \leftarrow \mathcal{A}_{gm}(t, G^{(t)}_e, E)\)
        \IF{\( Y_{t} == \text{False} \)}
            \STATE \( Y \leftarrow \text{False} \)
            \STATE \textbf{break}
        \ENDIF
        
    \ELSE
        \STATE \( e, Y_{t} \leftarrow \mathcal{A}_{gcp}(t, G^{(t)}_e, E) \)
        \IF{\( Y_{t} == \text{False} \)}
            \STATE \( Y \leftarrow \text{False} \)
            \STATE \textbf{break}
        \ELSE
            \STATE \( \mathcal{E}_c \leftarrow \mathcal{E}_c \cup e \)\COMMENT{Update Claim Graph}
            \STATE \( G_e \leftarrow G_e \cup \mathcal{A}_{gc}(E, [e]) \)\COMMENT{Update Evidence Graph}
        \ENDIF
    \ENDIF
\ENDFOR
\RETURN \( Y \)
\end{algorithmic}
\end{algorithm}

\section{Baseline Details}
\label{appendix:baselines}
We present detailed implementation for our baseline models across two categories:
\textbf{Pre-trained/Fine-tuned Models:}
\begin{itemize}
\item \textbf{BERT-FC} \cite{soleimani2020bert}: Employs \texttt{bert-large-uncased} (345M parameters) for binary classification, concatenating claims and evidence as '[CLS] claim [SEP] evidence'. Fine-tuned using 20 random examples from the dataset.
\item \textbf{LisT5} \cite{jiang2021exploring}: Implements \texttt{t5-large} with listwise concatenation, processing all evidence sentences simultaneously for binary classification (Supports/Refutes).
\item \textbf{RoBERTa-NLI} \cite{nie2019combining}: RoBERTa-large model fine-tuned on SNLI~\cite{DBLP:conf/emnlp/BowmanAPM15}, MNLI~\cite{DBLP:conf/naacl/WilliamsNB18}, FEVER-NLI~\cite{DBLP:conf/aaai/NieCB19}, ANLI (R1, R2, R3)~\cite{DBLP:conf/acl/NieWDBWK20} datasets, with additional fine-tuning using 20 task-specific examples from HOVER/FEVEROUS.
\item \textbf{DeBERTaV3-NLI} \cite{he2021debertav3}: DeBERTaV3-large model trained on 885,242 NLI pairs from FEVER and other NLI datasets (MNLI, ANLI, LingNLI~\cite{DBLP:conf/emnlp/ParrishHALNWAAL21}, WANLI~\cite{DBLP:journals/corr/abs-2201-05955}).
\item \textbf{MULTIVERS} \cite{wadden2021multivers}: Implements LongFormer \cite{beltagy2020longformer} architecture to handle extended evidence sequences, specifically fine-tuned on FEVER \cite{thorne2018fever} dataset.
\end{itemize}
\textbf{In-context Learning Models:}
\begin{itemize}
\item \textbf{Codex} \cite{chen2021evaluating}: Utilizes \texttt{code-davinci-002} with 20 in-context examples.
\item \textbf{FLAN-T5} \cite{chung2022scaling}: Leverages \texttt{FLAN-T5-XXL} (3B parameters) with 20 in-context examples, using structured prompts for claim verification.
\item \textbf{ProgramFC} \cite{pan2023fact}: Combines Codex and FLAN-T5 for generating reasoning programs. For fair comparison, we adopt their single reasoning chain configuration (N = 1).
\item \textbf{Direct, Decomposition}: Implements \texttt{gpt-3.5-turbo-0125} using two approaches: (1) zero-shot prompting (Direct) and (2) claim decomposition with 10 in-context examples (Decomposition). The latter approach first breaks down complex claims into independently verifiable textual sub-claims, then applies direct prompting to verify each sub-claim. The complete prompts are provided in Appendix \ref{appendix:prompts}. 
\end{itemize}

\section{Prompts}\label{appendix:prompts}
Below, we present the prompt templates used for the baseline methods (Direct, Decomposition) and {\MyFC}. For clarity, we have omitted the in-context-learning examples.
\begin{lstlisting}[caption=Direct Prompt, label={lst:direct_prompt}]
[[Evidence]]
Based on the above information, is it true that [[Claim]]? True or false? The answer is: 
\end{lstlisting}
\begin{lstlisting}[caption=Decompostion Prompt, label={lst:decompostion_prompt}]
## Task Description:
Break down the given into multiple sub claims that:
- Cannot be further divided into simpler meaningful statements
- Has a clear truth value (true or false)
- Contains a single subject and predicate
- Does not contain logical connectives (and, or, if-then, etc.)

## Input Format:
A complex claim or statement.

## Output Format:
Each atomic proposition should be presented on a new line, separated by blank lines for clarity.

## Examples:
(... more in-context examples here ...)

## Real Data:
Input: [[Claim]]
Output:
\end{lstlisting}
\begin{lstlisting}[caption=Claim Graph Construction Prompt, label={lst:claim_graph}]
# Knowledge Graph Construction Specification

You are an expert in knowledge graph construction. Your task is to parse natural language claims into a formal claim graph representation by following these specifications:

## 1. Entity Types
- Person: Real individuals
- Location: Places, cities, regions
- Organization: Companies, institutions, groups 
- Time: Dates, years, periods
- Number: Numerical values
- Concept: Abstract ideas, categories
- Object: Physical items
- Work: Creative works (books, songs, etc.)
- Event: Occurrences, happenings
- Species: Biological organisms

## 2. Constraint Types
- temporal: Time-related constraints (year, date, period)
- spatial: Location-related constraints (in, at, from)
- condition: Qualifying conditions or attributes
- context: Broader situational context

## 3. Atomic Proposition Rules

### Definition
An atomic proposition must:
- Express a single, indivisible fact
- Cannot be broken down into simpler meaningful statements
- Must preserve all relevant context
- Must maintain temporal and spatial relationships

### Decomposition Guidelines
1. Structural Analysis:
   - Split complex sentences at conjunction words (and, but, or)
   - Separate conditional statements (if/then) into distinct propositions
   - Identify dependent clauses and their relationships
   - Preserve modifiers and qualifiers with their related concepts

2. Semantic Preservation:
   - Maintain causal relationships
   - Preserve temporal order
   - Keep spatial relationships intact
   - Retain contextual qualifiers

## 4. Fuzzy Entity Identification Process

### Entity Reference Types
1. Direct Reference:
   - Uses proper name ("John", "Paris")
   - Specific numerical values
   - Well-defined concepts

2. Indirect Reference:
   - Uses descriptions ("the teacher", "that city")
   - Role-based references ("the founder", "the mother")
   - Attribute-based references ("the tall building")

3. Contextual Reference:
   - Requires information from other statements
   - Part of a collective reference
   - Implied entities

### Fuzzy Entity Decision Tree
1. Initial Check:
   - Is the entity referred to by proper name?  Not fuzzy
   - Is the entity a specific number or date?  Not fuzzy
   - Is the entity a well-defined concept?  Not fuzzy

2. Context Analysis:
   - Does the entity require contextual information?  Fuzzy
   - Is the entity part of a group or collection?  Fuzzy
   - Is the entity only described by role or attribute?  Fuzzy
   - Is the entity referenced through relationships?  Fuzzy

3. Coreference Resolution:
   - Track entities across multiple atomic propositions
   - Maintain consistent fuzzy entity IDs ($A$, $B$, etc.)
   - Document relationships between fuzzy entities

## 5. Output Format

### 1. Entity Definitions
```python
entities = [
    {
        "name": str,       # Entity identifier 
        "type": str,       # One of the predefined entity types
        "is fuzzy": bool   # True if entity meets fuzzy criteria
    }
]
```

### 2. Relationship Quadruples with Atomic Propositions
```python
quads = [
    {
        "atomic proposition": str,  # The original atomic statement this quad represents
        "subject": str,            # Entity name or identifier
        "predicate": str,          # Relationship type
        "object": str,             # Entity name, identifier, or value
        "constraint": {            # Optional constraint
            "type": str,           # One of the predefined constraint types
            "value": str           # Constraint value
        }
    }
]
```

## 6. Validation Steps

### 1. Quad and Atomic Proposition Validation
For each quad:
- Verify its atomic proposition cannot be further decomposed
- Check if it captures all relevant context
- Ensure temporal/spatial information is preserved
- Validate the relationship between entities
- Check if all relevant constraints are captured

### 2. Semantic Consistency Check
For the entire graph:
- Verify all quads together preserve the original claim's meaning
- Check for logical gaps or inconsistencies
- Validate temporal/causal order is maintained
- Ensure entity coreferences are consistent

## 7. Example 
(... more in-context examples here ...)

Now, please parse the following claim into the formal representation described above:
Claim = [[Claim]]

Output =

\end{lstlisting}
\begin{lstlisting}[caption=Evidence Graph Construction Prompt, label={lst:evidence_graph}]
## Task:  
Extract semantic triples from the given evidence and ensure every extracted triple is context-independent.  

## Input:  
- Evidence: [Full text of the evidence]  
- Entities: [List of entities mentioned, including entity name and its type]  

## Output Format:  
Extract triples separated by newlines, with each triple in the format:  
`Subject | Predicate | Object`  

## Extraction Guidelines:  
1. Ensure that either **Predicate** and **Object**, or both in each triple are from the provided Entity Set. 
2. "Identify all relationships" from the Evidence Text that meet this requirement.  
3. Every triple should be context-independent. Use full forms or expanded phrases for relational references where necessary. 

## Examples
(... more in-context examples here ...)

## Real data
Evidence: [[Evidence]]

Entity Set: 
[[Entity Set]]

Output:

\end{lstlisting}
\begin{lstlisting}[caption=Graph Match Prompt, label={lst:graph_match_prompt}]
Evidence:
[[Evidence]]  

Using the provided evidence, determine whether the given quadruple is true or false.  
Quadruple:
[[Quadruple]]  

Output(true/false): 
\end{lstlisting}
\begin{lstlisting}[caption=Graph Completion Prompt, label={lst:graph_completion_prompt}]
Evidence:
[[Evidence]]

Complete the fuzzy entity in the quadruple based on above evidence. If a suitable entity is found, output its name; otherwise, output "none".

Quadruples:
[[Quadruples]]

Output:
\end{lstlisting}

\end{document}